\documentclass[10pt,twocolumn,letterpaper]{article}

\usepackage{cvpr}              

\usepackage{times}
\usepackage{epsfig}
\usepackage{graphicx}
\usepackage{amsmath}
\usepackage{amssymb}
\usepackage{dsfont}
\usepackage{color}
\usepackage{xcolor}
\usepackage{comment}
\usepackage{mathabx}

\let\degree\relax      
\usepackage{gensymb}   
\usepackage{cite}
\usepackage[ruled,vlined]{algorithm2e}
\usepackage[format=plain,labelformat=simple,labelsep=period,font=small,compatibility=false]{caption}
\usepackage{subfigure}

\usepackage[pagebackref=true,breaklinks=true,letterpaper=true,colorlinks,bookmarks=false]{hyperref}

\usepackage[capitalize]{cleveref}
\crefname{section}{Sec.}{Secs.}
\Crefname{section}{Section}{Sections}
\Crefname{table}{Table}{Tables}
\crefname{table}{Tab.}{Tabs.}

\def\cvprPaperID{8159} 
\def\confName{CVPR}
\def\confYear{2022}

\definecolor{darkgreen}{RGB}{0,150,0}
\begin{document}

\title{Egocentric Scene Understanding via Multimodal Spatial Rectifier}

\author{Tien Do$^{1}$ \qquad \qquad Khiem Vuong$^{2}$ \qquad \qquad Hyun Soo Park$^{1}$\\\\
$^1$ University of Minnesota \qquad $^2$ Carnegie Mellon University}


\twocolumn[{%
\maketitle
	\begin{center}
		\centering
	 \includegraphics[width=1\linewidth]{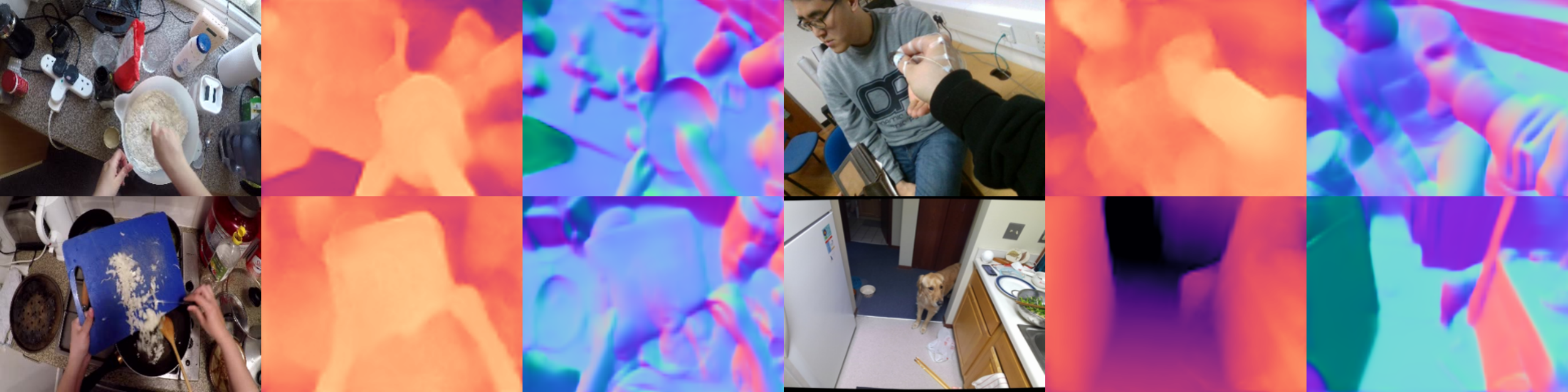}
	 \vspace{-6mm}
	\captionof{figure}{\small We study the problem of predicting geometry (depths and surface normals) from a single view egocentric image that includes dynamic objects (e.g., hand and people). We design a multimodal spatial rectifier that can effectively handle the excessively tilted images caused by head movement (e.g., nearly 90 degree pitch angle when engaging eye-hand coordination). Our method shows strong performance on unseen images from EPIC-KITCHENS~\cite{damen2018scaling} (left), FPHA~\cite{garcia2018first} (top right), and our EDINA (bottom right) datasets.} \label{fig:teaser}
\end{center}	
}]

\begin{abstract}
In this paper, we study a problem of egocentric scene understanding, i.e., predicting depths and surface normals from an egocentric image. Egocentric scene understanding poses unprecedented challenges: (1) due to large head movements, the images are taken from non-canonical viewpoints (i.e., tilted images) where existing models of geometry prediction do not apply; (2) dynamic foreground objects including hands constitute a large proportion of visual scenes.
These challenges limit the performance of the existing models learned from large indoor datasets, such as ScanNet~\cite{dai2017scannet} and NYUv2~\cite{Silberman:ECCV12}, which comprise predominantly upright images of static scenes.
We present a multimodal spatial rectifier that stabilizes the egocentric images to a set of reference directions, which allows learning a coherent visual representation. 
Unlike unimodal spatial rectifier that often produces excessive perspective warp for egocentric images, the multimodal spatial rectifier learns from multiple directions that can minimize the impact of the perspective warp.
To learn visual representations of the dynamic foreground objects, we present a new dataset called EDINA (Egocentric Depth on everyday INdoor Activities) that comprises more than 500K synchronized RGBD frames and gravity directions.
Equipped with the multimodal spatial rectifier and the EDINA dataset, our proposed method on single-view depth and surface normal estimation significantly outperforms the baselines not only on our EDINA dataset, but also on other popular egocentric datasets, such as First Person Hand Action (FPHA)~\cite{garcia2018first} and EPIC-KITCHENS~\cite{damen2018scaling}.
\end{abstract}

\section{Introduction}
We interact with surrounding objects in structured yet rather complex, unorganized, and dynamic environments, enabled by our robust egocentric perception that facilitates understanding 3D scene geometry around us. Such innate perceptual ability shows in stark contrast with that of existing computer vision systems, trained to operate on images depicting static and well-organized scenes recorded by carefully controlled cameras~\cite{dai2017scannet, Geiger2012KITTI, Silberman:ECCV12}. These trained models~\cite{fu2018dorn, Huang19FrameNet} are, despite their remarkable performance, shown to be highly brittle when predicting the scene geometry of egocentric images that observe unscripted everyday activities, including diverse hand-object interactions, captured by in situ embodied sensors such as head/body-mounted cameras~\cite{Do2020SurfaceNormal}. This requires additional sensors such as multi-camera rigs, IMU and depth sensors in augmented/mixed reality devices (e.g., Hololens and Magic Leap One) to deliver interactive and immersive experiences in our daily spaces.

In this paper, we study a problem of egocentric 3D scene understanding---predicting depths and surface normals from a single view egocentric image. In addition to challenges of classic scene understanding problems~\cite{dai2017scannet}, egocentric scene understanding poses two more challenges: (1) Images are no longer upright. Head movements induce significant roll and pitch motions where the scene is often depicted in a tilted way. In particular, by the nature of hand-eye coordination, egocentric images inherently are affected by severe pitch motion when manipulating objects, which is substantially different from the existing data distribution, e.g., ScanNet~\cite{dai2017scannet}, NYUv2~\cite{Silberman:ECCV12}, and KITTI~\cite{Geiger2012KITTI}.  (2) Images include not only background objects, e.g., furniture, room layout, and walls, but also  dynamic foreground objects, e.g., humans and arms/hands (see Figure~\ref{fig:teaser}). Classic scene understanding mainly focuses on reconstructing the overall geometric layout made of such background objects while the foreground ones are considered as outliers. In contrast, these foregrounds are more salient in egocentric scenes as they are highly indicative of evolving activities.

%
We conjecture that the challenges of egocentric scene understanding can be addressed by an image stabilization method that incorporates the fundamentals of equivariance, called spatial rectifier~\cite{Do2020SurfaceNormal}---an image warping that transforms a titled image to a canonical orientation (i.e., gravity-aligned) such that a prediction model can learn from the upright images. This is analogous to our robust perception through mental stabilization of visual stimuli~\cite{Wurtz2011Imagestabilization}. However, the spatial rectifier shows inferior performance on predicting 3D geometry of egocentric images that involve substantial head movement (e.g., nearly 90 degree pitch), leading to excessive perspective warps. 
We present a \textit{multimodal spatial rectifier} by generalizing the canonical direction, i.e., instead of unimodal gravity-aligned direction, we learn multiple reference directions from the orientations of the egocentric images, which allows minimizing the impact of excessive perspective warping.
Our multimodal spatial rectifier makes use the clusters of egocentric images based on the distribution of surface normals into multiple pitch modes, where we learn a geometric predictor (surface normals or depths) that is specialized for each mode to rectify associated roll angles. 

%

To facilitate learning the visual representation of dynamic egocentric scenes, we present a new dataset called \textit{EDINA} (Egocentric Depth on everyday INdoor Activities). 
Our dataset comprises more than 15 hours RGBD recording of indoor activities including cleaning, cooking, eating, and shopping. 
Our dataset provides a synchronized RGB, depth, surface normal, and the 3D gravity direction to train our multimodal spatial rectifier and geometry prediction models.
Our depth and surface normal predictor learned from the EDINA outperforms the baselines predictors not only on EDINA dataset but also other datasets, such as EPIC-KITCHENS~\cite{damen2018scaling} and First Person Hand Action (FPHA)~\cite{garcia2018first}.

Our contributions include: (1) a multimodal spatial rectifier; (2) a large dataset of egocentric RGBD with the gravity that is designed to study egocentric scene understanding, by capturing diverse daily activities in the presence of dynamic foreground objects; (3) comprehensive experiments to highlight the effectiveness of our multimodal spatial rectifier and our EDINA dataset towards depth and surface normal prediction on egocentric scenes.
\section{Related Works}
Our egocentric scene understanding lies in the intersection between single view geometry and equivariant spatial rectifier. We briefly review the related work.

\noindent\textbf{Single View Depth and Surface Normal} Single view scene understanding approaches have shown great progress by leveraging a large amount of data such as ScanNet~\cite{dai2017scannet} that supervises to learn the mapping from an image to a 3D scene geometry such as depths~\cite{Eigen2014depth, eigen2015DepthNormalSemanticLabel, Qi_2018_CVPR, chen2019structure, hao2018detail, hu2019revisiting, jiao2018look, laina2016deeper, lee2019monocular, ramamonjisoa2019sharpnet, ren2019deep, yin2019enforcing, zhang2019pattern,fu2018dorn, huynh2020guiding, Liu2019planercnn, Zhao_2021_CVPR_Depth_camera_pose, Ranftl2020MIDAS, Ranftl2021DPT} or surface normals~\cite{li2015depthnormalcrf, Chen2017SurfaceNormalsInTheWilds, eigen2015DepthNormalSemanticLabel,wang2016surge,bansal2016Marr2D3DalignmentViaSN,Qi_2018_CVPR,Huang19FrameNet,wang2020vplnet,zeng2019rgbdcompletion,Do2020SurfaceNormal}. 
Existing methods that show remarkable performance on scene understanding tasks have focused on either: (1) designing deep neural network architectures~\cite{fu2018dorn, Do2020SurfaceNormal, huynh2020guiding}; or (2) exploiting useful 2D visual cues for learning 3D geometry, including textures~\cite{Huang19FrameNet}, vanishing points~\cite{wang2020vplnet}, planar surfaces~\cite{wang2016surge, Liu2019planercnn}, and depth-surface normal consistency~\cite{Qi_2018_CVPR}.
Nevertheless, they are only enabled by large-scale indoor RGBD datasets, such as ScanNet~\cite{dai2017scannet}, NYUv2~\cite{Silberman:ECCV12}, Sun3D~\cite{xiao2013sun3d}, and Sun RGBD~\cite{song2015sun}.
%
%
However, due to the nature of data collection methods, a model trained on such datasets show a notable performance degradation when applying it to egocentric images because of two reasons: (1) The model has not exposed to the tilted images that have substantially different visual patterns from that of upright images. (2) The model has limited capability of learning dynamic foreground objects that are abundant in egocentric scenes, e.g., hands, pots, pans, vacuums, brooms, pets, and humans.
To address these challenges in egocentric images, we make use of a multimodal spatial rectifier, which allows using large existing datasets in conjunction with egocentric datasets.
%

\noindent\textbf{Rotation Equivariance} Equivariance is a geometric property of a visual representation: the visual representation in an image must be transformed, according to the transformation of the scene.
Enforcing equivariance in learning a scene geometry allows geometrically coherent learned models. 
To achieve this, camera poses~\cite{Zhao_2021_CVPR_Depth_camera_pose} and gravity directions~\cite{Do2020SurfaceNormal} can be employed.
For instance, equivariance is used to learn the geometry of scenes by augmenting transformations, i.e., spatial rectifier~\cite{Do2020SurfaceNormal} that rectifies a tilted image to an upright (gravity-aligned) image~\cite{saito2020roll_depth, Do2020SurfaceNormal, sartipi2020dde}.
Despite the substantial improvement on the tilted images, the spatial rectifier with a unimodal gravity-aligned direction shows poor performance on egocentric images.
It is mainly caused by excessive warping of egocentric images due to a large variation of camera angle, e.g., nearly 90 degree pitch angle when engaging eye-hand coordination.
Our multimodal spatial rectifier prevents such excessive perspective warp by predicting multiple reference directions, which significantly improves the egocentric scene understanding task.

\noindent\textbf{Egocentric Scene Datasets} Egocentric scene datasets have been used for a wide range of tasks such as action recognition~\cite{fathi2012learning,fathi2011understanding,pirsiavash2012detecting}, action anticipation~\cite{bertasius2018egocentric,rhinehart2017first}, and many others~\cite{furnari2016temporal,furnari2018personal,furnari2016recognizing}. 
Notably, Damen et al.~\cite{damen2018scaling} proposed EPIC-KITCHENS, a large-scale egocentric benchmark with densely annotated actions and object interactions in the kitchen environment. 
%
%
A few egocentric RGBD datasets that exist were designed for activity recognition~\cite{garcia2018first, moghimi2014experiments, rogez2015understanding, tang2018multi}. 
With a few exception, such datasets do not include the 3D gravity direction that is critical for learning an equivariant representation. 
Our EDINA dataset provides synchronized RGBD and gravity directions captured from an egocentric viewpoint with diverse daily activities. 
%
\section{Method}

We present a multimodal spatial rectifier that stabilizes tilted images into multiple transformation modes. This method minimizes the impact of perspective warping while retaining equivariance property.

\begin{figure}[t]
\centering
\includegraphics[width=0.4\textwidth]{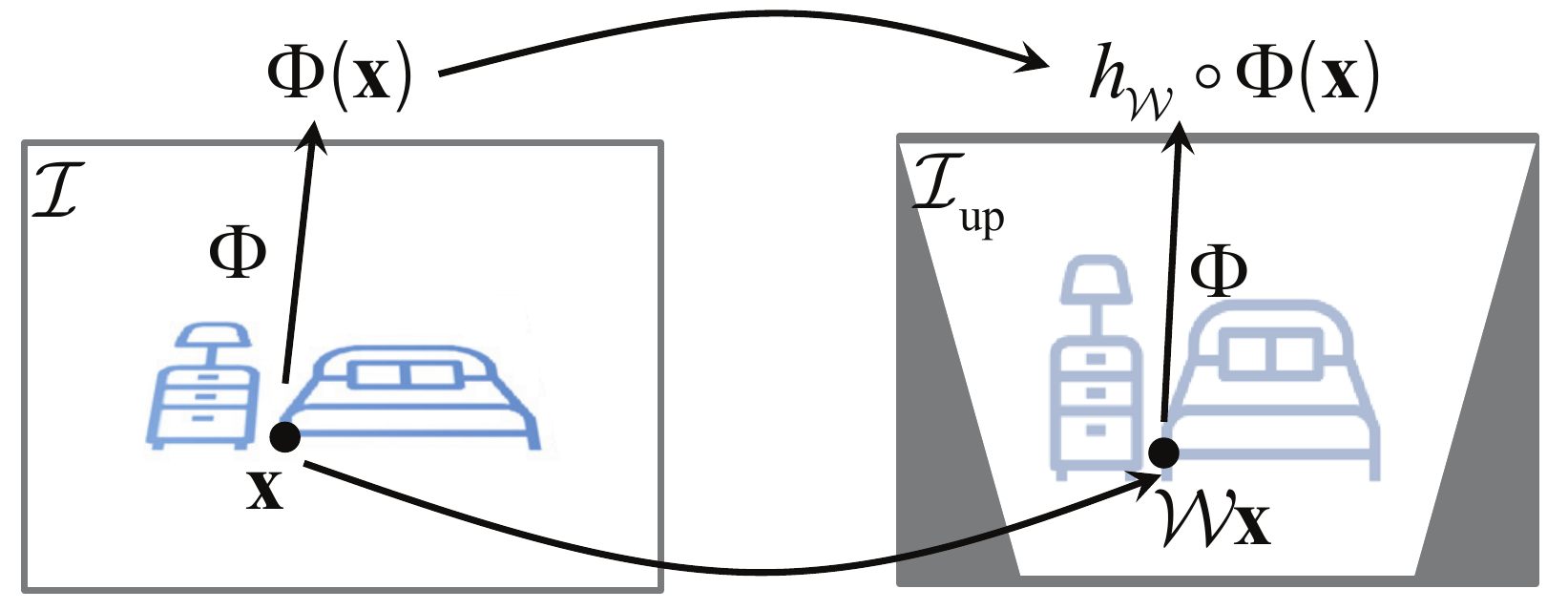}
\caption{A spatial rectifier enforces equivariance property to learn a geometrically coherent representation. When a point is transformed by $\mathcal{W}$, its feature is expected to transformed accordingly, i.e., $\Phi(\mathcal{W}\mathbf{x}) = h_{\mathcal{W}}\circ \Phi(\mathbf{x})$.
}\vspace{-5mm}
\label{fig:equivariance}
\end{figure}

\subsection{Equivariant Spatial Rectifier}
\label{subsection:prelimsr}

Consider a function $\Phi:\mathds{R}^2\times \mathds{I}\rightarrow \mathds{R}^n$ that predicts the geometry of a pixel $\mathbf{x}\in\mathds{R}^2$ in an image $\mathcal{I} \in \mathds{I}$, where  $\mathds{I}=[0,1]^{\scriptsize{3\times H \times W}}$ is the image range ($H$ and $W$ are its height and width, respectively).
We denote the prediction:
\begin{align}
    y = \Phi(\mathbf{x}, \mathcal{I}),
\end{align}
where $y \in \mathds{R}^n$ and $n$ is the dimension of the geometry, e.g., $n=1$ for depth, and $n=3$ for surface normal. 

\begin{figure}[t]
    \begin{center}
    \includegraphics[width=0.49\textwidth]{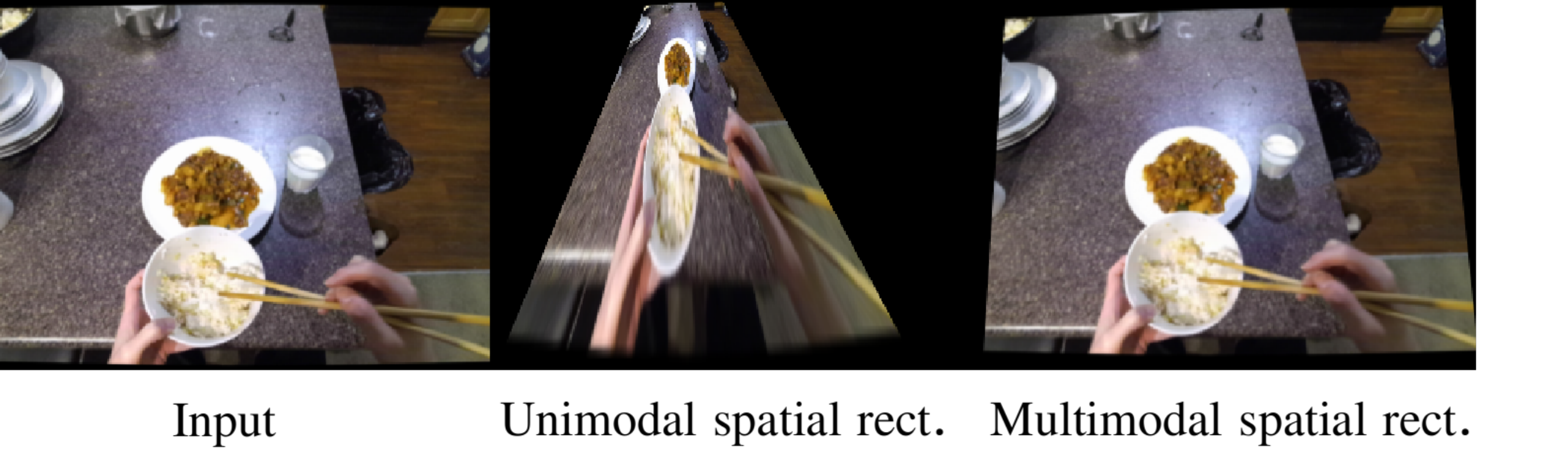}
    \end{center}
    \vspace{-7mm}
    \caption{A unimodal spatial rectifier produces an excessive perspective warp (middle) to align the image to the gravity direction, which significantly degrade the performance of geometry prediction. We use a multimodal spatial rectifier that warps to multiple reference directions that minimizes the impact of the perspective warping (right).}\vspace{-5mm}
    \label{fig:excessive_warp}
\end{figure}

A spatial rectifier~\cite{Do2020SurfaceNormal} is learned to transform a \textit{tilted} image $\mathcal{I}$ with the gravity direction $\mathbf{g}\in \mathds{S}^2$ in the camera coordinate system to the \textit{upright} image $\mathcal{I}_{\rm up}$ with the upright gravity direction $\mathbf{g}_{\rm up}$ by explicitly enforcing an equivariant property through 3D rotation (Figure~\ref{fig:equivariance}):
\begin{align}
    h_\mathcal{W} \circ \Phi(\mathbf{x}, \mathcal{I}) =  \Phi(\mathcal{W}(\mathbf{x}; \mathbf{R}_{\rm up}), \mathcal{I}_{\rm up}),
\end{align}
where $\mathcal{W}:\mathds{R}^2\times SO(3)\rightarrow\mathds{R}^2$ is a 2D transformation that maps a point in the tilted image to the upright image based on the 3D gravity direction. That is, the transformation can be determined by a homography induced by camera pure rotation $\mathbf{R}_{\rm up}\in SO(3)$ such that $\mathbf{g}_{\rm up} = \mathbf{R}_{\rm up}\mathbf{g}$. 
$\mathcal{I}_{\rm up}$ is warped from the tilted image by $\mathcal{W}$, i.e., $\mathcal{I}_{\rm up} = \mathcal{I}(\mathcal{W}(\mathbf{x}; \mathbf{R}_{\rm up}))$. $h_\mathcal{W}$ is the geometry transformation parametrized by $\mathcal{W}$, e.g., (1) for the surface normal prediction, $h_\mathcal{W}$ is equivalent to rotating the surface normal vector ($\mathds{S}^2$), i.e., $h_\mathcal{W}\circ \Phi = \mathbf{R}_{\rm up}\Phi$; (2) for the depth prediction, $h_\mathcal{W}$ is defined as:
\begin{align}
    h_\mathcal{W}\circ \Phi = \left(\mathbf{R}_{\rm up}\mathbf{K}^{-1}\widetilde{\mathbf{x}}\right)_{\rm z}\Phi
\end{align}
where $(\mathbf{v})_z$ denote the $3^{\text{rd}}$ coordinate of a vector $\mathbf{v}\in\mathds{R}^{3}$, and $\mathbf{K}$ is the camera intrinsic matrix, and $\widetilde{\mathbf{x}} \in \mathds{P}^2$ is the homogeneous representation of $\mathbf{x}$.

\begin{figure*}[t]
\centering
\includegraphics[width=\textwidth]{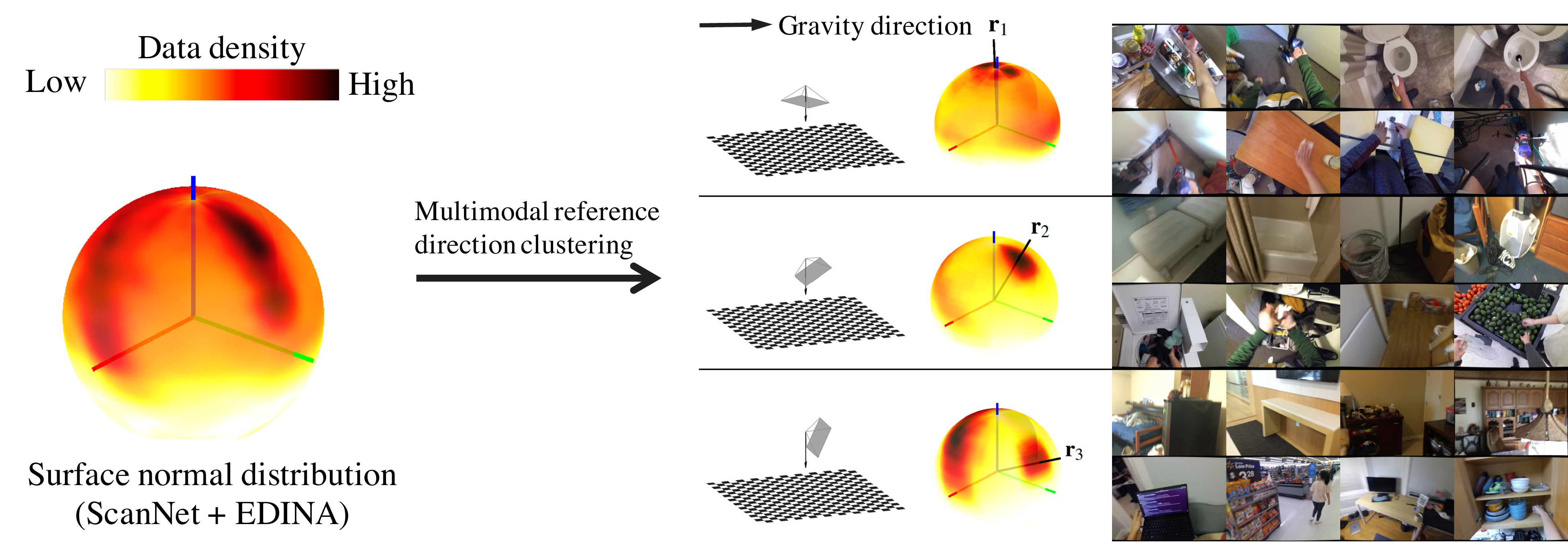}
\vspace{-7mm}
\caption{Unlike the spatial rectifier~\cite{Do2020SurfaceNormal} that relies on the unimodal surface normal distribution with respect to the gravity direction (left), we present a multimodal spatial rectifier that generalizes the spatial rectifier by learning multiple reference directions (right). As a result, the surface normal distribution of the scene datasets can be decomposed into multiple clusters, which allows minimizing the impact of image warping and more importantly, learning a geometrically coherent representation.  
}\vspace{-5mm}
\label{fig:edina_cluster}
\end{figure*}

Predicting the geometry of a tilted image can be modeled as a function composition:
\begin{align}
     \Phi(\mathbf{x}, \mathcal{I}) = h_\mathcal{W}^{-1} \circ \Phi_{\rm up}(\mathcal{W}(\mathbf{x}; \mathbf{R}_{\rm up}), \mathcal{I}_{\rm up}),\label{Eq:sr}
\end{align}
where $h_\mathcal{W}^{-1}$ is the spatial rectifier, and $\Phi_{\rm up}$ is the geometry predictor learned from upright images. A key benefit of this function composition is that $\Phi_{\rm up}$ can be trained solely by the large training dataset made of the upright images (e.g., ScanNet~\cite{dai2017scannet} and NYUv2~\cite{Silberman:ECCV12}), which can be, in turn, used to predict the surface normals of a tilted image. 

\noindent\textbf{Limitation} Despite of its strong performance on tilted images, the spatial rectifier exhibits a major limitation towards egocentric scene understanding due to its single modal rectification. 
The spatial rectifier is designed to warp a tilted image with respect to a single upright direction, which applies to roll and mild pitch camera rotations. 
In constrast, egocentric images often have substantial head orientation due to the hand-eye coordination, resulting in severe perspective warped image $\mathcal{I}_{\rm up}$ (e.g., 90$^{\circ}$ pitch tilted image), which in turns, significantly degrades the performance of the geometry predictor as shown in Figure~\ref{fig:excessive_warp} (middle).

\subsection{Multimodal Spatial Rectifier}
\label{subsection:msr}

We generalize the spatial rectifier model by leveraging a mixture of expert models~\cite{masoudnia2014MixtureExperts} called \textit{multimodal spatial rectifier} where each expert model predicts the spatial rectification:
\begin{align}
    \Phi(\mathbf{x},\mathcal{I}) = \frac{1}{\sum_i b_i}\sum_i b_i \left(h_{\mathcal{W}_i}^{-1} \circ \Phi_i(\mathcal{W}(\mathbf{x}; \mathbf{R}_i), \mathcal{I}_i)\right),\label{eq:msr}
\end{align}
where $b_i\in\mathds{R}_{+}$ is a non-negative weight to mix transformations, and $\mathbf{R}_i$ is the rotation that transforms the gravity of the tilted image to the $i^{\rm th}$ reference direction, i.e., $\mathbf{r}_i = \mathbf{R}_i \mathbf{g}$. $\mathcal{I}_i$ is warped from the tilted image by $\mathcal{W}_i$, i.e., $\mathcal{I}_i = \mathcal{I}(\mathcal{W}(\mathbf{x}; \mathbf{R}_i))$. The reference direction $\mathbf{r}\in \mathds{S}^2$ is a generalization of the upright gravity $\mathbf{g}_{\rm up}$, which specifies the egocentric tilted images to be warped. $\Phi_i$ is the geometry predictor designed for the $i^{\rm th}$ reference direction. We denote $\mathcal{W}(\mathbf{x}; \mathbf{R}_i)$ by $\mathcal{W}_i$ by abuse of notation.
The key benefit of the multimodal spatial rectifier is the flexibility of image warping. The severe head orientation of an egocentric image can be warped to the closest reference direction, which prevents excessive perspective warping (see Figure~\ref{fig:excessive_warp}).

We find the set of reference directions $\{\mathbf{r}_i\}_{i=1}^K$ along the pitch directions by clustering the gravity directions of egocentric images with $K$ is the predefined number of the reference directions:
\begin{align}
    \underset{\{\mathbf{r}_i\}_{i=1}^K}{\text{minimize}}  \sum_{i=1}^{K} \sum_{j \in \mathcal{C}_i} ~~ \|\mathbf{g}_j - \mathbf{r}_i\|^{2}_{2}, \label{eq:cluster_gravity}
\end{align}
where $\mathcal{C}_i$ is the set of the indices of training instances of which gravity directions closest to the $i^{\rm th}$ reference direction $\mathbf{r}_i$. 
%
In practice, we design an iterative algorithm inspired by K-Medoids algorithm~\cite{Park2009Kmedoids} by increasing the number of cluster numbers $K$ until the total deviation reaches below a threshold $\delta$ indicating the data is well-fitted (see Algorithm~\ref{alg:dataclustering}).
\begin{algorithm}[b]
\SetAlgoLined
\SetKwInOut{Input}{Input}
\SetKwInOut{Output}{Output}
    \Input{$\delta, \{ \mathbf{g}_j \}_{ \mathcal{I}_j \in \mathcal{D}_{\rm train} }$}
    \Output{\{$\mathbf{r}_{i}\}_{i=1}^{K}$}
 $K=1, t=\delta+\epsilon$\;
 \While{$t > \delta$}{
  $\{\mathbf{r}_{i}\}_{i=1}^{K}$ = K-Medoids$\left(\{\mathbf{g}_j\}_{\mathcal{D}_{\rm train}}, K\right)$\;
  $t = \sum_{i=1}^{K} \sum_{j \in \mathcal{C}_i} ~ \|\mathbf{g}_j - \mathbf{r}_i\|^{2}_{2}$\;
  $K \gets K+1$\;
 }
 \caption{Determine reference directions} \label{alg:dataclustering}
\end{algorithm}
Figure~\ref{fig:edina_cluster} illustrates gravity cluster centers and images as well as their surface normal map belonging to each cluster. Similar to spatial rectifier~\cite{Do2020SurfaceNormal}, we represent a 3D rotation by two unit vectors: ($\mathbf{g}$, $\mathbf{e}$) are gravity and principle direction. $\mathbf{e}$ is the unit vector that is a mode of distribution that represents surface normals in an image (see details in Appendix). In practice, we use one-hot encoding for $\{b_i\}$, i.e., $b_i=1$ if $\mathbf{r}_i$ is closest to $\mathbf{g}$, and zero otherwise. 
\vspace{-2mm}
\subsection{Learning Spatial Rectifier}
\label{subsection:training}
\vspace{-2mm}
We learn a spatial rectifier given a set of ground truth directions $\{(\mathcal{I},\mathbf{g}, \mathbf{e}, \mathbf{y})\}_{\mathcal{D}}$ where $\mathcal{D}$ is the training dataset. 
$\mathbf{y}\in \mathds{R}^{n\times H\times W}$ is the ground truth geometry ($n=1$ for depth and $n=3$ for surface normal). 

Consider two learnable functions $f_{\mathbf{g}}, f_\mathbf{e}: \mathds{I}\rightarrow \mathds{S}^2$ that predict the gravity and principle directions from an image, respectively. 
These two functions constitute a spatial rectifier that can be learned by minimizing the following loss: 
\begin{equation}
    \mathcal{L}_{\rm SR}(\mathcal{I}, \mathbf{g}, \mathbf{e}) =  \text{cos}^{-1}(\mathbf{g}^{\mathsf{T}} f_{\mathbf{g}}(\mathcal{I})) + \text{cos}^{-1}(\mathbf{e}^{\mathsf{T}} f_{\mathbf{e}}(\mathcal{I})),
\end{equation}
%
%

We jointly learn the multimodal spatial rectifier together with the geometry predictor by minimizing the following loss:
\begin{align}
    \mathcal{L} = \sum_{\{\mathcal{I}, \mathbf{g}, \mathbf{e}, \mathbf{y}\}\in\mathcal{D}} \mathcal{L}_{\rm GEO}(\mathbf{y}, \mathcal{I}) + \lambda \mathcal{L}_{\rm SR}(\mathcal{I}, \mathbf{g}, \mathbf{e}). \label{eq:geometric_loss}
\end{align}
%
%
The geometric loss $\mathcal{L}_{\rm GEO}$ measures the geometric error between the prediction and ground truth:
\begin{align}
\mathcal{L}_{\rm GEO}(\mathbf{y}, \mathcal{I}) &= \sum_{\mathbf{x}} d(\mathbf{y}_{\mathbf{x}}, \boldsymbol\Phi(\mathbf{x}, \mathcal{I})),  \text{ where }\nonumber  \\
d(y, \Phi) &= \begin{cases}
    |y-\Phi| & {\text{ for~depth}} \\
    \text{cos}^{-1}\left(y^{\mathsf{T}}\Phi\right)& {\text{ for~surface~normal}}
\end{cases}\nonumber
\end{align}
where $\boldsymbol\Phi(\mathbf{x}, \mathcal{I})=h^{-1}_{\mathcal{W}}\circ\boldsymbol\Phi(\mathcal{W}(\mathbf{x}; \mathbf{R}), \overline{\mathcal{I}})$, and $\mathbf{R}$ can be computed by the predictions of $f_\mathbf{g}(\mathcal{I})$ and $f_\mathbf{e}(\mathcal{I})$.
\begin{figure}[t]
  \centering  
      \includegraphics[width=0.49\textwidth]{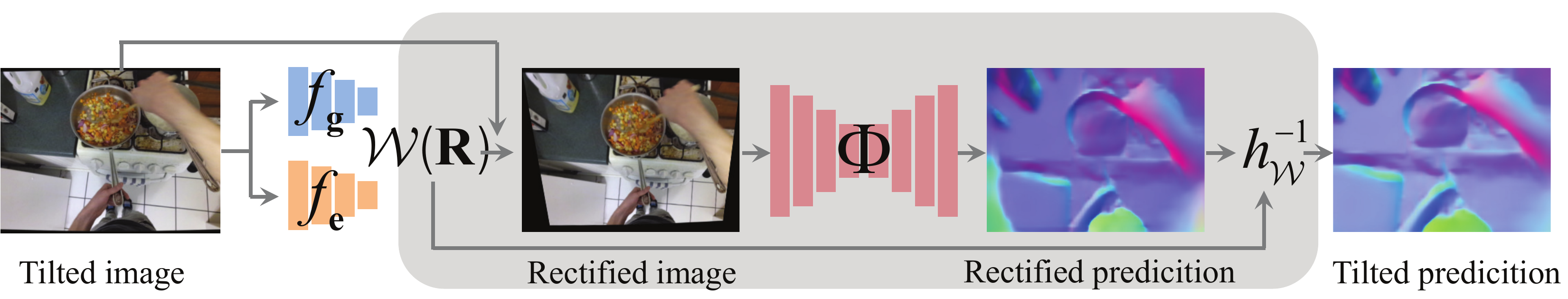}
    \vspace{-5mm}
  \caption{The multimodal spatial rectifier warps an egocentric image by predicting the gravity $\mathbf{g}$ and principle directions $\mathbf{e}$s, allowing learning a coherent geometry predictor $\Phi$.} \label{fig:multimodal_sr} \vspace{-5mm}
\end{figure}

\subsection{Network Design}
The multimodal spatial rectifier is a modular predictor that can combine with a geometry predictor $\Phi$ as shown in Figure~\ref{fig:multimodal_sr}. It is learned to predict the gravity and principle directions from an input tilted image through $f_{\mathbf{g}}$ and $f_{\mathbf{e}}$, respectively. With the predicted direction, it computes the rotation $\mathbf{R}$ that can be used to warp the image to the reference direction $\mathcal{W}$. The geometry predictor takes as input an image and predict depths and surface normals. These predictions are unwarped by $h^{-1}_\mathcal{W}$. 

\noindent \textbf{Implementation Details} Our networks take as input an RGB image of size $320\times240$ and output the same size surface normals or depths. We use a ResNet-18 architecture to estimate $f_{\mathbf{g}}$ and $f_{\mathbf{e}}$ while the geometry predictor $\Phi$ is specified in~\ref{sec:expr_baselines}. The proposed models are implemented in PyTorch~\cite{pytorch}, trained with a batch size of 32 on a single NVIDIA Tesla V100 GPU, and optimized by Adam~\cite{kingma1412adam} optimizer with a learning rate of $10^{-4}$. We train our models for 20 epochs.



\section{EDINA Dataset}

\begin{figure*}[t]
  \centering  
      \subfigure[EDINA image, depth, surface normal, and gravity direction]{\label{fig:dataset}\includegraphics[width=\textwidth]{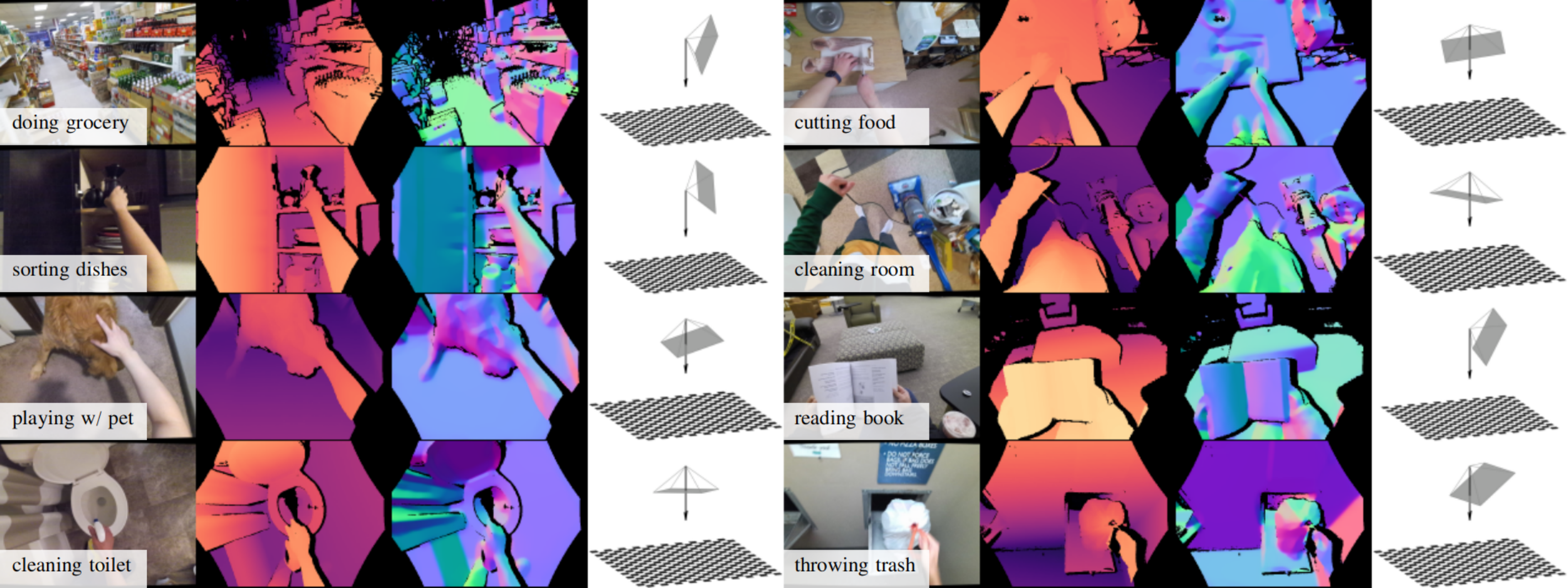}}\\
      \subfigure[Gravity distribution]{\label{fig:edina_gravity_distribution}\includegraphics[height=0.25\textheight]{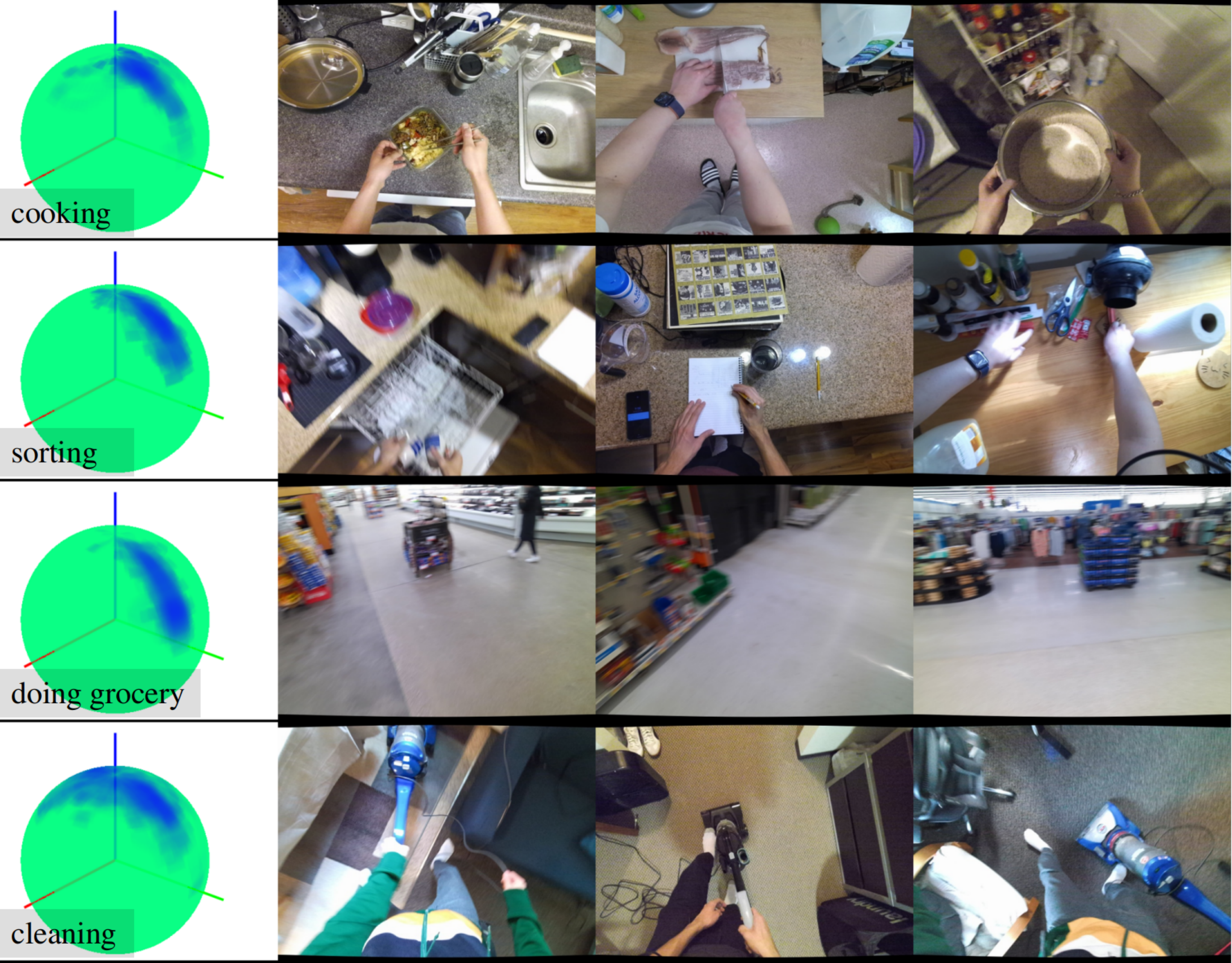}}~~~~~~~~
      \subfigure[Activities]{\label{Fig:activity}\includegraphics[height=0.23\textheight]{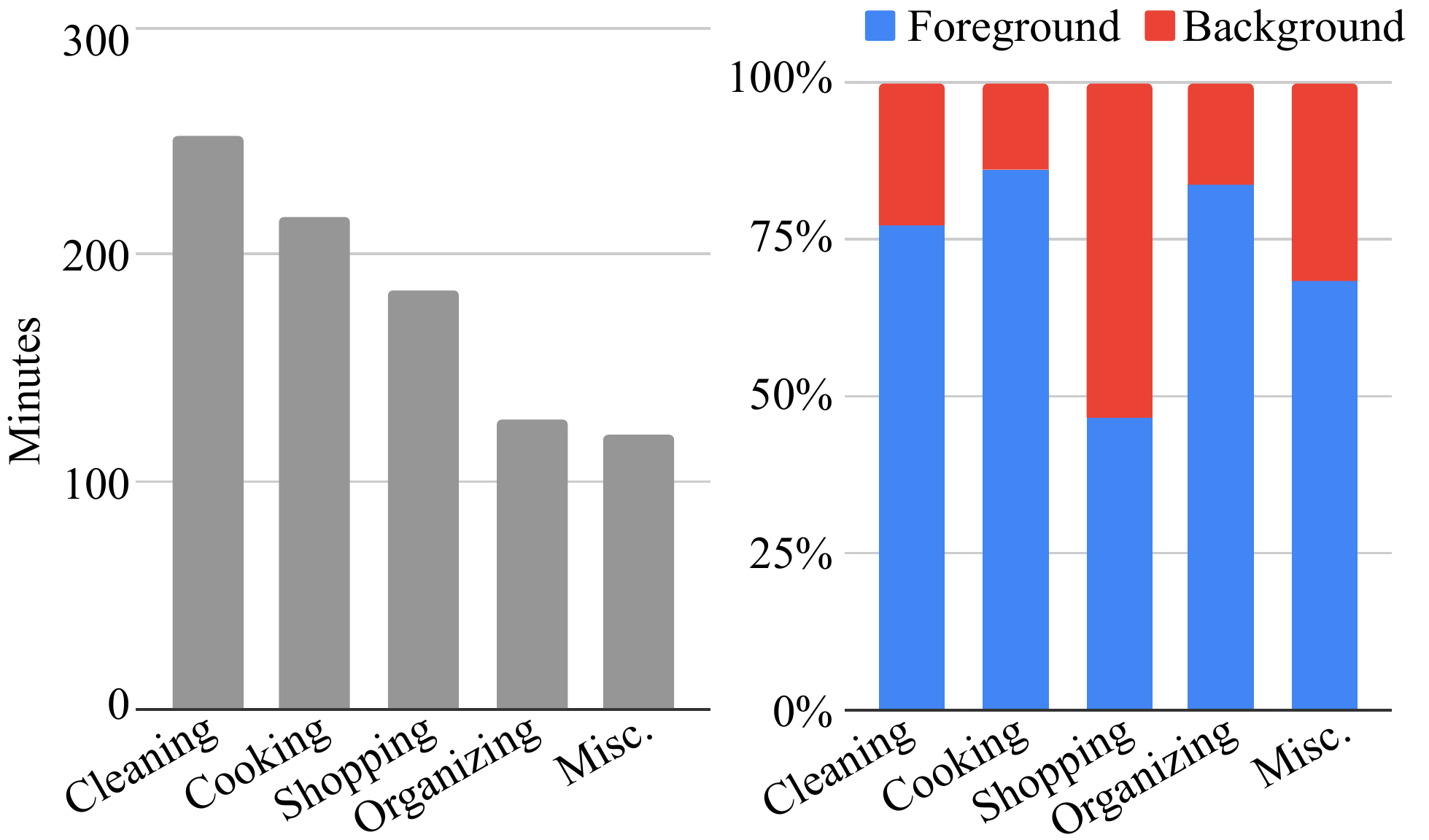}}
  \caption{We present EDINA (Egocentric Depth on everyday INdoor Activities) dataset. (a) We show egocentric images of diverse activities with depths, surface normals, and gravity direction (black). (b) Gravity direction is highly correlated with egocentric activities. The images of cooking and cleaning activities have nearly 90$^\circ$ pitch angle, which is different from shopping activities. (c) EDINA includes four major indoor activities of cleaning, cooking, shopping, and home organizing. Unlike existing scene datasets such as ScanNet, a large proportion of pixels of egocentric scenes belong to the foreground.} \label{Fig:dacc}
\end{figure*}

We present a new RGBD dataset called \textit{EDINA} (Egocentric Depth on everyday INdoor Activities) that facilitates learning 3D geometry from egocentric images. Each instance in the dataset is a triplet: RGB image (1920$\times$1080), depths and surface normals (960$\times$540), and 3D gravity direction. The data were collected using Azure Kinect cameras~\cite{KinectAzure} that provide RGBD images (depth range: 0.5$\sim$5.46m) with inertial measurement unit signals. Eighteen participants were asked to perform diverse daily indoor activities, e.g., cleaning, sorting, cooking, eating, doing laundry, training/playing with pet, walking, shopping, vacuuming, making bed, exercising, throwing trash, watering plants, sweeping, wiping, while wearing a head-mounted camera. The camera is oriented to approximately 45$^\circ$ downward to ensure observing hand-object interactions. Total number of data instances is 550K images (16 hrs). Figure~\ref{fig:dataset} illustrates the representative examples of EDINA dataset that include substantially tilted egocentric images depicting diverse activities.  


The gravity direction is highly correlated with activities. For instance, the majority of cooking and cleaning activities are performed while facing down, whereas the shopping and interacting with others are performed while facing front as shown in Figure~\ref{fig:edina_gravity_distribution}. Figure~\ref{Fig:activity} illustrates the amount of data of four major indoor activities of cleaning, cooking, shopping, and home organizing. Unlike existing scene datasets such as ScanNet, a large proportion of pixels of egocentric scenes belong to the foreground.
%
%
Our dataset is available at \url{https://github.com/tien-d/EgoDepthNormal}.

\section{Experiments}

We evaluate our two main contributions: accuracy of multimodal spatial rectifier and effectiveness on multiple datasets including EDINA.

\begin{figure}[t]
\centering
\includegraphics[width=0.38\textwidth]{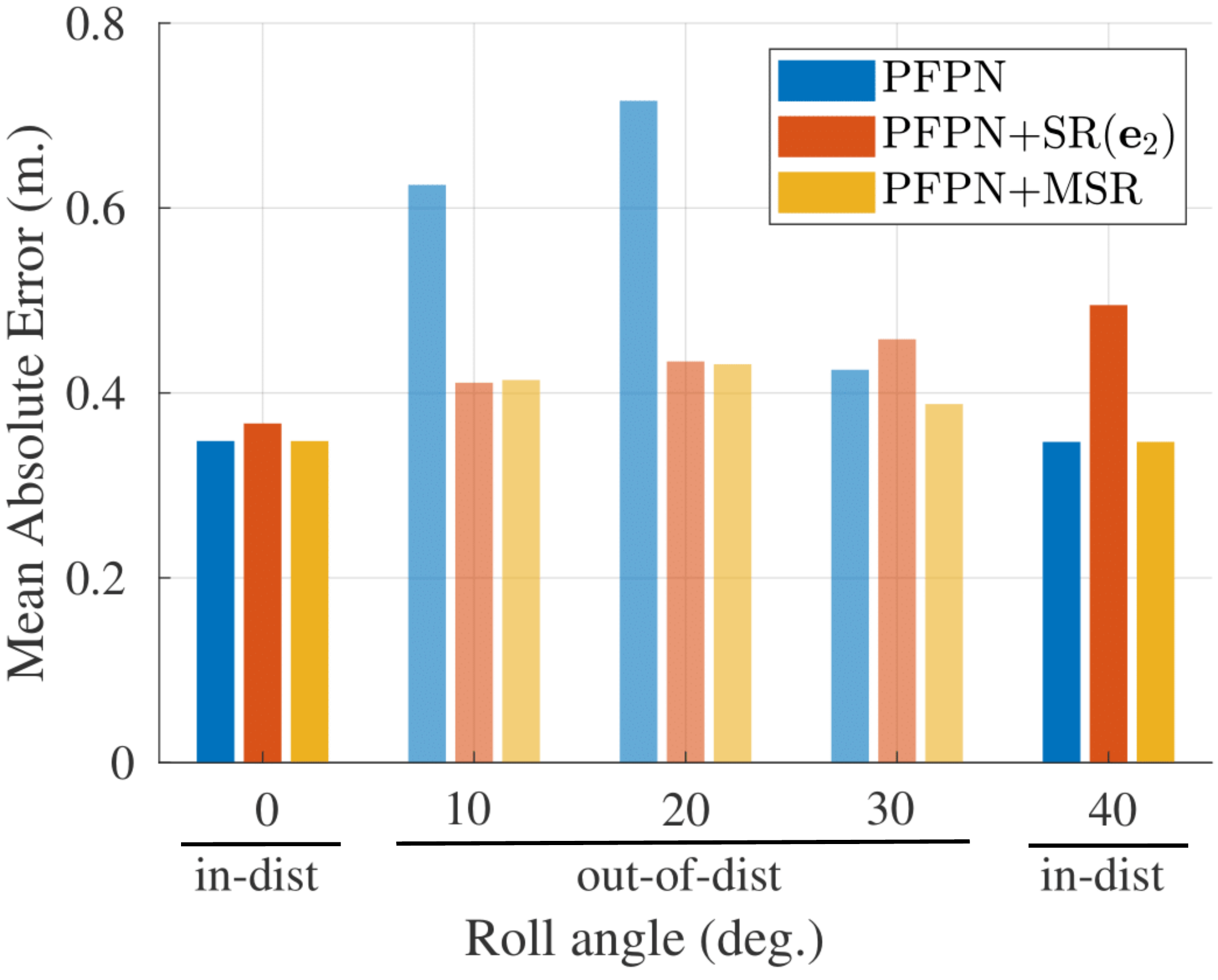}
\vspace{-3mm}
\caption{Performance of PFPN, PFPN+SR($\mathbf{e}_2$), and PFPN+MSR on HM3D test set. The dark and light color indicates the in- (at 0${}^{\circ}$ and 40${}^{\circ}$) and out-of-distribution (at 10${}^{\circ}$, 20${}^{\circ}$, 30${}^{\circ}$), respectively.} 
\vspace{-6mm}
\label{fig:sanity_check}
\end{figure}

\subsection{Evaluation Datasets}

\noindent\textbf{HM3D}~\cite{Ramakrishnan:2021} To facilitate more controlled experiments, we use HM3D, a large-scale dataset containing 1,000 distinctive building-scale, real-world 3D reconstructions. The data are composed of textured 3D mesh reconstruction with high visual fidelity, which allows us to render the photo-realistic scenes from diverse viewpoints with known camera orientations. We render the RGB-D frames from each viewpoint and only retain the views that are complete (no missing surfaces or reconstruction artifacts). \noindent\textbf{ScanNet~\cite{dai2017scannet}} ScanNet is a large RGB-D indoor datasets with 1500 sequences, spanning a wide variety of scenes. We use the standard dataset split used in FrameNet~\cite{Huang19FrameNet} that comprises 199,720 frames for training and 64,319 frames for validating. In addition, we utilize FrameNet's high-quality ground-truth surface normals to augment with our EDINA for training.\\ \begin{table*}[t]
\begin{center}
\resizebox{\textwidth}{!}{
\begin{tabular}{l|l|cccc|ccc}
\hline
 Testing & Method & Abs. Rel$\downarrow$ & Sq. Rel$\downarrow$ & log-RMSE$\downarrow$ &  
 RMSE $\downarrow$ &	1.25$\uparrow$ &	$1.25^2 \uparrow$ &	$1.25^3 \uparrow$ \\
\hline
& MiDaS (\texttt{MIX6})${}^{\dagger}$ & 0.194  & 0.079  &  0.267  & 0.247  & 68.20 &  83.96 &  93.14 \\
& DPT (\texttt{MIX6})${}^{\dagger}$ & 0.195  & 0.073  &  0.256  & 0.234  & 66.95 &  86.07 &  94.39 \\
& PFPN (\texttt{ScanNet}) & 0.536 & 0.292 & 0.450 & 0.410 & 28.50 & 63.31 & 84.60 \\
EDINA & PFPN (\texttt{EDINA}) & 0.173	& 0.052 & 0.210	& 0.181	& 78.81	& 92.97	& 97.06 \\
& PFPN & 0.161 &	0.044 & 0.197 & 0.168 & 81.03 & 94.16 & 97.68 \\
& PFPN+MSR (Ours) & \textbf{0.145} ({\footnotesize \textcolor{darkgreen}{-9.7\%}}) &	\textbf{0.041} ({\footnotesize \textcolor{darkgreen}{-8.5\%}}) &	\textbf{0.182} ({\footnotesize \textcolor{darkgreen}{-7.7\%}}) &		\textbf{0.155} ({\footnotesize \textcolor{darkgreen}{-7.9\%}}) & \textbf{84.06} & \textbf{94.54} & \textbf{97.87} \\
\hline
& PFPN (\texttt{ScanNet}) & 1.252 &	0.893	& 0.788	&  0.580	& 10.36	 & 28.07	 & 48.87 \\
FPHA & PFPN (\texttt{EDINA}) & 1.229	& 4.114	& 0.802		& 1.483	& 25.98	& 46.38	& 62.70 \\
& PFPN & 0.737	& 0.457	 & 0.549		& 0.397	 & 32.60	& 57.61	& 75.14 \\
& PFPN+MSR (Ours) & \textbf{0.657}	({\footnotesize \textcolor{darkgreen}{-10.8\%}}) & \textbf{0.369}	({\footnotesize \textcolor{darkgreen}{-19.2\%}}) & \textbf{0.508}	({\footnotesize \textcolor{darkgreen}{-7.3\%}}) & \textbf{0.337}	({\footnotesize \textcolor{darkgreen}{-15.2\%}}) & \textbf{37.70}	& \textbf{62.50}	& \textbf{78.30} \\
\hline
\end{tabular}}
\caption{We compare the performance of depth prediction of our method (MSR) with baselines on EDINA and FPHA testing data. The ${}^{\dagger}$ indicates methods that predict scale-ambiguous depth and thus require a scale correction step. The numbers in the parenthesis show the percentage of the reduction in error metrics of PFPN+MSR (Ours) with respect to the baseline PFPN, where the \textcolor{darkgreen}{green} highlight denote this improvement in percentage.}
\vspace{-6mm}
\label{tab:depth_edina}
\end{center} \vspace{-3mm}
\end{table*}
\noindent \textbf{Evaluation Metrics} We assess the accuracy of the predicted depths using multiple standard metrics, including: (a) mean absolute relative error (Abs. Rel), (b) mean square relative error (Sq. Rel), (c) logarithmic root mean square error (log-RMSE), (f) root mean square error (RMSE), and (g) the percentage of the estimated depths $\hat{d}$ for which  $\max(\frac{\hat{d}}{d^{*}},\frac{d^{*}}{\hat{d}}) < \delta$, where $d^{*}$ is the ground-truth depth and $\delta=1.25,1.25^2,1.25^3$.
%
In terms of surface normal error metrics, we also employ standard metrics originally used in~\cite{bansal2016Marr2D3DalignmentViaSN,Fouhey_2013_ICCV}: (a) mean absolute of the error (Mean), (b) median of absolute error (Median), (c) root mean square error (RMSE), and (d) the percentage of pixels with angular error below a threshold $\xi$ with $\xi = $ 5\degree, 7.5\degree, 11.25\degree.
\noindent\textbf{EDINA (ours)} We use EDINA dataset to train and evaluate our models on surface normal and depth estimation. With a total of 550K RGB-D images and IMU measurements, we include 500K images collected by 15 participants in the training set and use the remaining 50K images collected by the rest of the three participants as the testing set. We also follow the approach of~\cite{Ladicky2014DiscriminativelySurfaceNormal} to generate ground truth surface normals from the depth images.
\noindent\textbf{FPHA~\cite{garcia2018first}} We use FPHA that is an egocentric RGB-D dataset consisting of 1,175 video sequences in several different hand-action categories for a total of 105,459 RGB-D frames and follow its official train/test split.
\subsection{Baselines}\label{sec:expr_baselines}
We construct various baseline algorithms using the state-of-the-art scene understanding approaches. (1) PFPN: Panoptic FPN~\cite{Kirillov19PanopticFPN} is a lightweight network architecture which has been used in various high-resolution prediction tasks. We employ PFPN with the ResNet-101~\cite{He_2016_CVPR} backbone as our baseline network architecture for both depth and surface normal estimation tasks.
(2)~PFPN+SR($\mathbf{e}_2$): we train PFPN using the spatial rectifier~\cite{Do2020SurfaceNormal} (SR) with a unimodal reference direction $\mathbf{e}_2 = \begin{bmatrix}0 \ 1 \ 0\end{bmatrix}^\mathsf{T}$.
(3)~PFPN+SR($\mathbf{e}_3$): we train PFPN using the SR with a unimodal reference direction $\mathbf{e}_3 = \begin{bmatrix}0 \ 0 \ 1\end{bmatrix}^\mathsf{T}$.
(4) PFPN+MSR: we train PFPN with our multimodal spatial rectifer (MSR) described in Section~\ref{subsection:msr}.
(5-8) DORN: DORN~\cite{fu2018dorn} is a high-capacity network architecture that is recently utilized in state-of-the-art surface normal estimation methods~\cite{Huang19FrameNet, Do2020SurfaceNormal}. Similar to PFPN, we also train DORN with the unimodal spatial rectifier on two reference directions $\mathbf{e}_2$, $\mathbf{e}_3$ and with our multimodal spatial rectifier, denoted by DORN+SR($\mathbf{e}_2$), DORN+SR($\mathbf{e}_3$), and DORN+MSR, respectively.
(9) MiDaS~\cite{Ranftl2020MIDAS}, (10) DPT~\cite{Ranftl2021DPT}: state-of-the-art depth prediction model trained on a large scale dataset MIX6~\cite{Ranftl2021DPT}. 
Since the depth prediction from MiDaS and DPT is ambiguous up to a scale factor, we scale the predicted depth maps with a common factor computed from the least-squares method~\cite{Xian_2018_mono_rel_depth, chen2016_single_depth_in_the_wild} using the ground truth depth on the train set.
We denote a network trained on a dataset by METHOD (\texttt{DATASET}), e.g., PFPN (\texttt{EDINA}) is the PFPN network that is trained on EDINA dataset. By default, all networks are trained on \texttt{ScanNet+EDINA}.
\begin{figure*}[t]
\includegraphics[width=\textwidth]{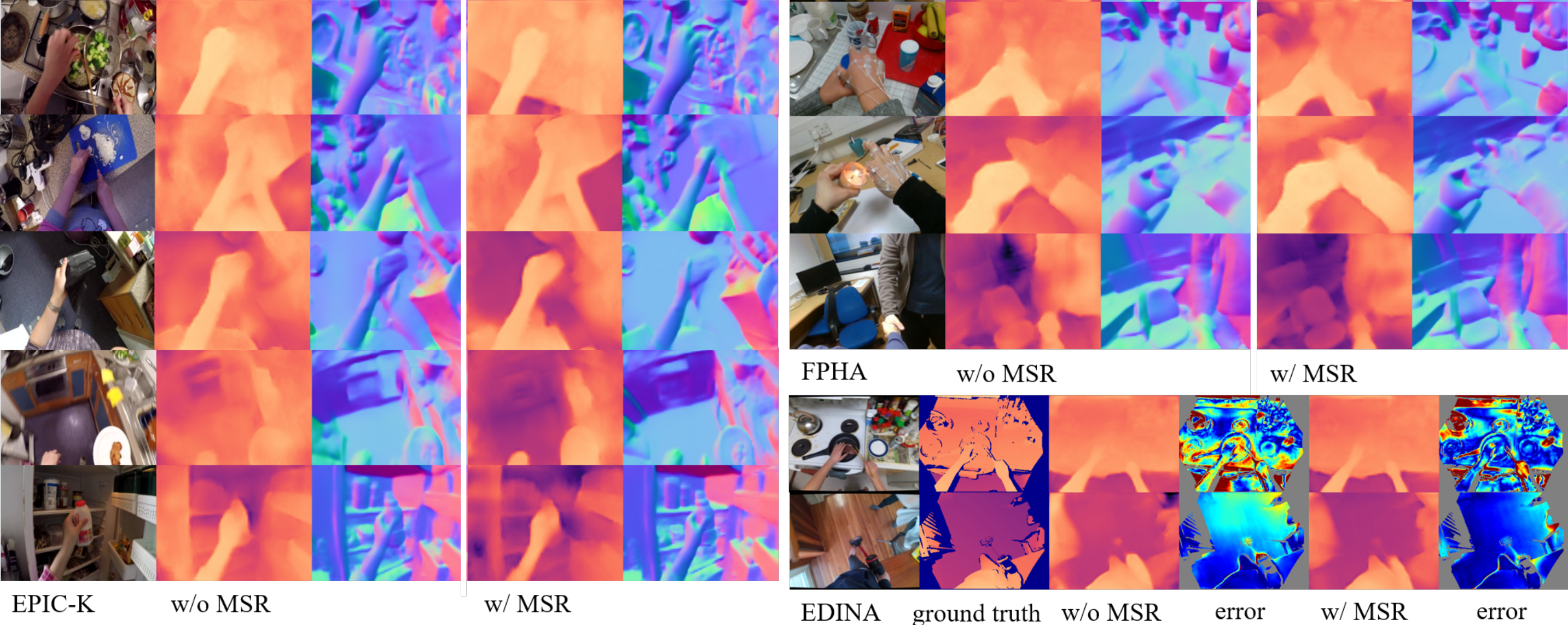}
\centering
\vspace{-6mm}
\caption{Qualitative results for EPIC-KITCHENS (Left), FPHA (top right), and EDINA (bottom right). For EPIC-KITCHENS and FPHA, from left to right: (1)~RGB image, (2)~depths and surface normals using PFPN trained on ScanNet and Edina, and (3)~depths and surface normals using PFPN+MSR trained on ScanNet and Edina. For EDINA, from left to right: (1)~RGB image, (2)~ground truth depths, (3)~estimated depths (w/o and w/ MSR), (4)~the corresponding depth error (w/o and w/ MSR).} \label{fig:qualitative}\vspace{-5mm}
\label{fig:result}
\end{figure*}

\subsection{Performance Benchmark}
\noindent\textbf{Depth Prediction} We first show the effectiveness of our MSR through a controlled experiment using the HM3D dataset.
Specifically, we render from HM3D a training set containing 82,941 RGB-D frames respectively at upright (tilt 0${}^{\circ}$) and tilt 40${}^{\circ}$ orientation and a testing set containing 3,944 RGB-D frames respectively at upright and tilt angles at 10${}^{\circ}$, 20${}^{\circ}$, 30${}^{\circ}$, and 40${}^{\circ}$.
The tilted images are rendered with the rotation around $\mathbf{e}_3$ axis with respect to the upright orientation (roll).
Figure~\ref{fig:sanity_check} illustrates the performance between PFPN, PFPN+SR($\mathbf{e}_2$), and PFPN+MSR (at two distribution modes 0${}^{\circ}$ and 40${}^{\circ}$) in 2 cases: (i) in-distribution: 0${}^{\circ}$ and 40${}^{\circ}$, and (ii) out-of-distribution: 10${}^{\circ}$, 20${}^{\circ}$, and 30${}^{\circ}$.
We can observe that for the in-distribution case, the baseline and MSR performs similarly while the PFPN+SR($\mathbf{e}_2$) slightly underperforms the former ones due to its excessive warping.
On the other hand, for the out-of-distribution case, while the baseline method degenerates at 10${}^{\circ}$, 20${}^{\circ}$, and 30${}^{\circ}$, both SR and MSR generalize reasonably well with the SR slightly degenerates when the tilt angle is further from its central mode (upright).

Table~\ref{tab:depth_edina} demonstrates the performance of our multimodal spatial rectifier and the effectiveness of our EDINA dataset.
A baseline network equipped with our spatial rectifier (PFPN+MSR) outperforms other baselines on nearly all evaluation metrics, not only on our EDINA dataset but also on FPHA dataset.
While the performance margin for the network equipped with and without MSR is narrow on EDINA, it is significant when generalizing to FPHA.
We conjecture that EDINA dataset that comprises a large variation in pitch angles can be overfitted by a large capacity network such as PFPN.
In contrast, FPHA dataset is taken from a shoulder-mounted camera, imposing more roll motion on the image, thus it causes a strong degradation for PFPN trained on ScanNet+EDINA datasets.
We conclude that our MSR module is highly beneficial for learning egocentric scene geometry.
Figure~\ref{fig:qualitative} illustrates the qualitative results of our method on EPIC-KITCHENS and FPHA. More qualitative results can be found in Supplementary Materials.

In addition, baselines that do not employ egocentric data, i.e., MiDaS (\texttt{MIX6}), DPT (\texttt{MIX6}), PFPN (\texttt{ScanNet}), performs poorly on both EDINA and FPHA.
On the other hand, the network trained only on EDINA performs strongly on its own test set while lacking generalizability towards to other dataset such as FPHA.
This indicates that learning can greatly benefit from a large amount of high quality ground truth geometry from ScanNet, together with our EDINA.

\noindent\textbf{Surface Normal Prediction}
In Table~\ref{tab:normal_edina}, we compare our method with the baselines on EDINA dataset and demonstrate the effectiveness of our proposed multimodal spatial rectifier on surface normal prediction.
On median and tight thresholds ($\xi = $ 5\degree, 7.5\degree), the unimodal spatial rectifier with $\mathbf{e}_2$ as the reference direction (PFPN+SR ($\mathbf{e}_2$)) shows notable improvements compared to the baseline PFPN while inferior in terms of RMSE and mean. Moreover, this issue further escalates when $\mathbf{e}_3$ is used as the only reference direction (PFPN+SR ($\mathbf{e}_3$)). This is mainly caused by the excessive warping that is very common on egocentric data.
In contrast, by predicting the multiple reference directions, our PFPN+MSR can generalize to diverse viewpoints, thus outperforms other baselines on all  metrics. 
Note that this also applies for DORN+MSR, suggesting that it is highly flexible and can be easily integrated into other networks.
See Figure~\ref{fig:result} for qualitative results.
\begin{table}[t]
\begin{center}
\resizebox{\columnwidth}{!}{
\begin{tabular}{l|cccccc}
\hline
 Method & Mean$\downarrow$ & Median$\downarrow$ & RMSE$\downarrow$ & 5\degree$\uparrow$ & 7.5\degree$\uparrow$ &11.25\degree$\uparrow$ \\
\hline
PFPN & 20.24 &	13.61 &	27.51 &	15.46 &	26.93 &	42.63 \\
PFPN+SR ($\mathbf{e}_2$) & 20.27 &	13.41 &	28.47 &	25.10 &	34.00 &	44.81 \\
PFPN+SR ($\mathbf{e}_3$) & 39.20 &	31.19 &	50.63 &	16.29 &	23.47 &	30.06 \\
PFPN+MSR & \textbf{19.30} &	\textbf{12.54} &	\textbf{27.37} &	\textbf{26.00} &	\textbf{35.49} &	\textbf{46.74} \\
\hline
DORN & 19.57 &	12.92 &	27.07 &	17.42 &	29.01 &	44.66 \\
DORN+SR ($\mathbf{e}_2$) & 19.96 &	12.68 &	28.46 &	25.53 &	35.00 &	46.35 \\
DORN+SR ($\mathbf{e}_3$) & 21.99 &	14.83 &	30.46 &	21.33 &	29.83 &	40.87 \\
DORN+MSR & \textbf{18.56} &	\textbf{11.55} &	\textbf{26.83} &	\textbf{26.58} &	\textbf{37.04} &	\textbf{49.18} \\
\hline
\end{tabular}}
\vspace{-3mm}
\caption{We compare the performance of surface normal prediction of our method (MSR) with baselines including the unimodal spatial rectifier (SR) on EDINA testing data.}
\label{tab:normal_edina}
\vspace{-8mm}
\end{center}
\end{table}

\section{Summary}
In this paper, we present a new multimodal spatial rectifier for egocentric scene understanding, i.e., predicting depths and surface normals from a single view egocentric image.
The multimodal spatial rectifier identifies multiple reference directions to learn a geometrically coherent representation from tilted egocentric images. 
This rectifier enables warping the image to the closest mode such that the geometry predictor in this mode can accurately estimate the geometry of the rectified scene.
To facilitate the learning of our multimodal spatial rectifier, we introduce a new dataset called EDINA that comprises 550K synchronized RGBD and gravity data of diverse indoor activities.
We show that EDINA is complementary to ScanNet, allowing us to learn a strong multimodal spatial rectifier.
We evaluate our method on egocentric datasets including our EDINA, FPHA and EPIC-KITCHENS, which outperforms the baselines.
%

\noindent \textbf{Acknowledgements} This work is partially supported by NSF CAREER IIS-1846031.

{\small
\bibliographystyle{ieee_fullname}
\bibliography{main}
}

\appendix
\begin{table*}[t]
	\begin{center}
		\resizebox{\textwidth}{!}{
			\begin{tabular}{l|l|cccc|ccc}
				\hline
				Testing & Method & Abs. Rel$\downarrow$ & Sq. Rel$\downarrow$ & log-RMSE$\downarrow$ & RMSE $\downarrow$ &	1.25$\uparrow$ &	$1.25^2 \uparrow$ &	$1.25^3 \uparrow$ \\
				\hline
				& PFPN (\texttt{THU-READ}) & 0.405 &	0.210 &	1.044 & 0.431 &  28.71 & 45.72 & 61.97 \\
				& PFPN (\texttt{FPHA}) & 0.314 &	0.167 &	0.500 &	0.378 &  42.82 & 66.48 & 78.98 \\
				& PFPN (\texttt{ScanNet}) & 0.536 & 0.292 & 0.450 & 0.410 & 28.50 & 63.31 & 84.60 \\
				& MiDaS (\texttt{MIX6})${}^{\dagger}$ & 0.194  & 0.079  &  0.267  & 0.247  & 68.20 &  83.96 &  93.14 \\
				& DPT (\texttt{MIX6})${}^{\dagger}$ & 0.195  & 0.073  &  0.256  & 0.234  & 66.95 &  86.07 &  94.39 \\
				\cline{2-9}
				& PFPN (\texttt{EDINA}) & 0.173	& 0.052 & 0.210	& 0.181	& 78.81	& 92.97	& 97.06 \\
				EDINA & PFPN & 0.161 &	0.044 & 0.197 & 0.168 & 81.03 & 94.16 & 97.68 \\
				& PFPN+SR($\mathbf{e_2}$) & 1.573 & 3.145	& 0.938 & 1.155  & 5.58 & 19.75 & 42.32 \\
				& PFPN+SR($\mathbf{e_3}$) & 0.381 & 0.333 & 0.475 & 0.416 & 51.75 & 73.37 & 84.62 \\
				& PFPN+MSR (Ours) & \textbf{0.145} ({\footnotesize \textcolor{darkgreen}{-9.7\%}}) &	\textbf{0.041} ({\footnotesize \textcolor{darkgreen}{-8.5\%}}) &	\textbf{0.182} ({\footnotesize \textcolor{darkgreen}{-7.7\%}}) &		\textbf{0.155} ({\footnotesize \textcolor{darkgreen}{-7.9\%}}) & \textbf{84.06} & \textbf{94.54} & \textbf{97.87}	 \\
				\hline
				& PFPN (\texttt{THU-READ}) & 0.439  & 0.150  &  4.629  & \textbf{0.279}  & 32.03 &  54.92 &  70.76 \\
				& PFPN (\texttt{ScanNet}) & 1.252 &	0.893	& 0.788	 & 0.580	& 10.36	 & 28.07	 & 48.87 \\
				& PFPN (\texttt{EDINA}) & 1.229	& 4.114	& 0.802	& 1.483	& 25.98	& 46.38	& 62.70 \\
				FPHA & PFPN & 0.737	& 0.457	 & 0.549		& 0.397	 & 32.60	& 57.61	& 75.14 \\
				& PFPN+MSR (Ours) & \textbf{0.657}	({\footnotesize \textcolor{darkgreen}{-10.8\%}}) & \textbf{0.369}	({\footnotesize \textcolor{darkgreen}{-19.2\%}}) & \textbf{0.508}	({\footnotesize \textcolor{darkgreen}{-7.3\%}}) & 0.337	({\footnotesize \textcolor{darkgreen}{-15.2\%}}) & \textbf{37.70}	& \textbf{62.50}	& \textbf{78.30} \\
				\cline{2-9}
				& PFPN (\texttt{FPHA}) & 0.119  & 0.023  &  0.139  & 0.075  & 91.29 &  97.31 &  98.75 \\
				\hline
		\end{tabular}}
		\caption{We compare the performance of depth prediction of our method (MSR) with baselines on EDINA and FPHA testing data. The ${}^{\dagger}$ indicates methods that predict scale-ambiguous depth and thus require a scale correction step. The numbers in the parenthesis show the percentage of the reduction in error metrics of PFPN+MSR (Ours) with respect to the baseline PFPN, where the \textcolor{darkgreen}{green} highlight denote this improvement in percentage.}
		\label{tab:depth_edina}
	\end{center}
\end{table*}

\section{Computing Principle Direction $\mathbf{e}$}

We describe the procedure to compute the principle direction $\mathbf{e}$ that are used in Equation (7) in the main manuscript.
For the $j^{\rm th}$ image, we find the weight $b_{ji}$ using the hard assignment by restricting $b_{ji} =\{0,1\}$ where $b_{ji}$ is one if $\mathbf{r}_i$ is closest to $\mathbf{g}$, and zero otherwise.
After specifying a cluster of egocentric images based on the reference direction, we find the surface normal distribution per cluster:
\begin{align}
	Q_i = \frac{1}{|\mathcal{C}_i|}\sum_{j\in \mathcal{C}_i} \mathtt{hist}(\mathbf{n}_j),\label{Eq:Qk}
\end{align}
where $\mathtt{hist}(\mathbf{n}_j)$ is the angular histogram of the surface normals of the $j^{\rm th}$ training image. $\mathbf{n}_j$ is the $3\times n$ matrix that each column presents a pixel's surface normal direction and $n$ is the total pixel in image $\mathcal{I}_j$.
The optimal rotation for the $j^{\rm th}$ image towards the $i^{\rm th}$ reference direction, $\mathbf{R}_{ji}^*$, is the one that maximizes the similarity in the surface normal distributions:
\begin{align}
	\mathbf{R}_{ji}^* = \underset{\mathbf{R}_{ji}}{\operatorname{argmin}}~D_{\rm KL}
	\left(\mathtt{hist}(\mathbf{R}_{ji}\mathbf{n}_j)||{Q}_i\right), \label{eq:klobj}
\end{align}
where $\mathbf{R}_{ji}\mathbf{n}_j$ is the rotated surface normals, and $D_{\rm KL}$ is KL divergence~\cite{Kullback:1951}. 
We optimize Equation~\eqref{eq:klobj} with an initial guess of $\mathbf{R}_{ji}$ computed by the gravity and reference directions:
\begin{align}
	\mathbf{R}_{ji} = \mathbf{I}_{3} + 2\mathbf{r}_i{\mathbf{g}}_j^\mathsf{T} 
	- \frac{\left(\mathbf{r}_i + \mathbf{g}_j\right)\left(\mathbf{r}_i+\mathbf{g}_j\right)^\mathsf{T}}{1 +\mathbf{r}_i^\mathsf{T}\mathbf{g}_j} ,\label{eq:rot_matrix_gravity_alignment}
\end{align}
where $\mathbf{I}_{3}$ is the $3\times 3$ identity matrix.

This optimal rotation can be parametrized by the principle direction $\mathbf{e}_j$~\cite{Do2020SurfaceNormal}, where $\mathbf{e}$ can be computed by:
\begin{equation}
	\mathbf{e}_j=\mathbf{R}_{ji}^*\mathbf{g}_j.\label{eq:e}
\end{equation}
We use the optimal principle direction as a ground truth to learn the multimodal spatial rectifier.
%

\section{Hardware setup}
\begin{figure}[t]
	\includegraphics[width=0.9\columnwidth]{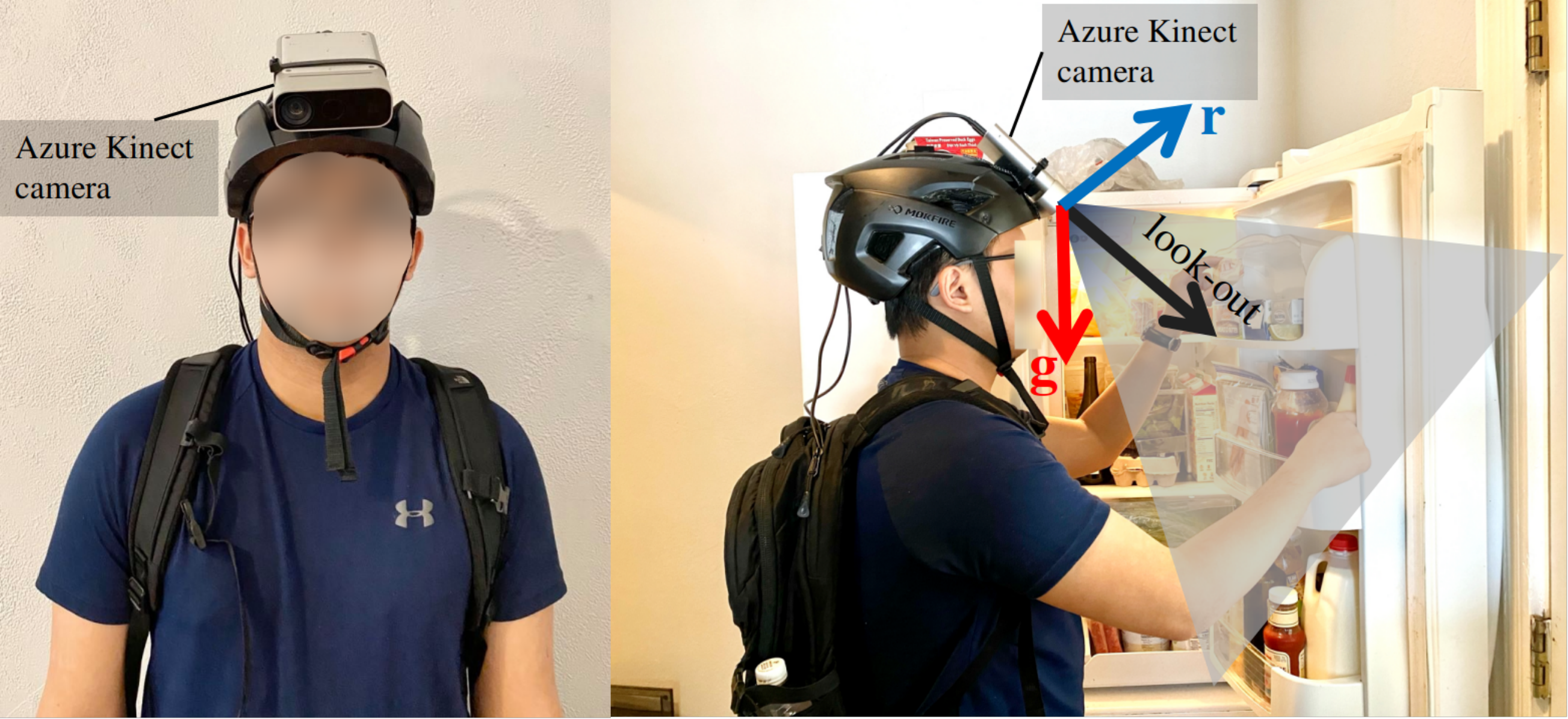}
	\centering
	\caption{EDINA dataset recording setup.} 
	\label{fig:camera_config}
\end{figure}

The participants were asked to wear an Azure-Kinect-mounted helmet while performing diverse daily indoor activities. The sensor was also connected to a laptop which reads and stores the raw data from the Azure Kinect device using the provided SDK. Figure~\ref{fig:camera_config} demonstrates the mounting configuration in which the camera is oriented to approximately 45$^\circ$ downward so that the captured interactions are within the field-of-view of the camera.

\section{More Results}

\noindent \textbf{Baselines} In addition to the datasets mentioned in the main manuscript, we also perform experiments on THU-READ~\cite{tang2017thuread}. THU-READ is an egocentric RGB-D dataset consisting of 1,920 video sequences in several different hand-action categories for a total of 171,474 RGB-D frames. We follow THU-READ's official 3/1 split for training/testing.

\noindent \textbf{Evaluation Metrics} We assess the accuracy of the predicted depths using multiple standard metrics, including: (a) mean absolute relative error (Abs. Rel), (b) mean square relative error (Sq. Rel), (c) logarithmic root mean square error (log-RMSE), (f) root mean square error (RMSE), and (g) the percentage of the estimated depths $\hat{d}$ for which  $\max(\frac{\hat{d}}{d^{*}},\frac{d^{*}}{\hat{d}}) < \delta$, where $d^{*}$ is the ground-truth depth and $\delta=1.25,1.25^2,1.25^3$.

\begin{figure}[h]
	\centering
	\vspace{-5mm}
	\includegraphics[width=0.3\textwidth]{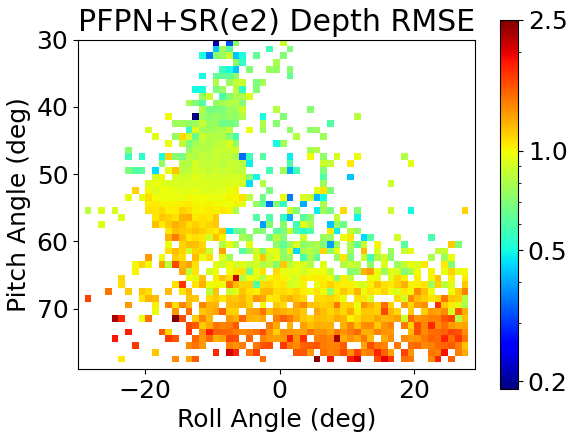}
	\vspace{-4mm}
	\hspace{-20pt}
	\caption{\small PFPN+SR($e_2$) error map w.r.t. different roll and pitch angles (on a subset of test images).}
	\vspace{-6mm}
	\label{fig:inandout}
\end{figure}

\noindent\textbf{Depth Prediction}
Table~\ref{tab:depth_edina} summarizes the performance of our multimodal spatial rectifier and the effectiveness of our EDINA dataset.
A baseline network with our multimodal spatial rectifier (PFPN+MSR) outperforms other baselines on nearly all evaluation metrics, not only on our EDINA dataset but also on FPHA dataset.
%
%
%
We conjecture that EDINA dataset that comprises a large variation in pitch angles can be overfitted by a large capacity network such as PFPN.
In addition, due to this substantial roll and pitch angles, it results in significant performance degradation for SR(${\mathbf{e}}_2$) or SR(${\mathbf{e}}_3$) (which motivates our multimodal spatial rectifier). 
This is also shown in the performance SR(${\mathbf{e}}_2$) on depth RMSE with respect to camera roll and pitch angle in Figure~\ref{fig:inandout}.
Furthermore, the FPHA dataset is taken from a shoulder mounted camera, imposing more roll motion on the image, thus it causes a strong degradation for PFPN trained on ScanNet+EDINA datasets.
We conclude that our MSR module is highly beneficial for learning egocentric scene geometry.
%

A baseline PFPN trained on THU-READ, FPHA, and ScanNet performs poorly on EDINA.
In addition, the baseline PFPN trained on THU-READ tends to generalize relatively well on FPHA because both datasets include hand-object interactions.
On the other hand, the network trained only on EDINA performs strongly on its own test set while lacking generalizability towards to other dataset such as FPHA.
Our baseline PFPN trained on ScanNet and EDINA outperforms PFPN trained on other datasets on FPHA.
This indicates that learning can greatly benefit from a large amount of high quality ground truth geometry from ScanNet, together with our EDINA.

\begin{figure}[h]
		\vspace{-3mm}
		\begin{center}
			\includegraphics[width=0.37\textwidth]{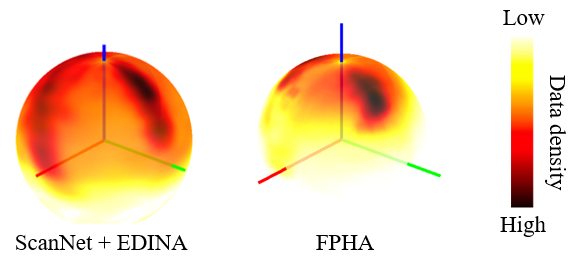}
		\end{center}
		\vspace{-7mm}
		\caption{\small ScanNet+EDINA vs FPHA surface normal distribution.}
		\label{fig:fpha_sphere}
		\vspace{-3mm}
	\end{figure}

\noindent {\bf Comparison of the clustered reference distributions of different datasets.} We show in Figure~\ref{fig:fpha_sphere} the comparison between ScanNet+EDINA and FPHA surface normal distribution. Note that FPHA has stronger distribution on the tilted modes due to the shoulder mounted camera, which shows the strong generalization capacity of our proposed MSR.

\begin{figure*}[b]
		\includegraphics[width=0.95\textwidth]{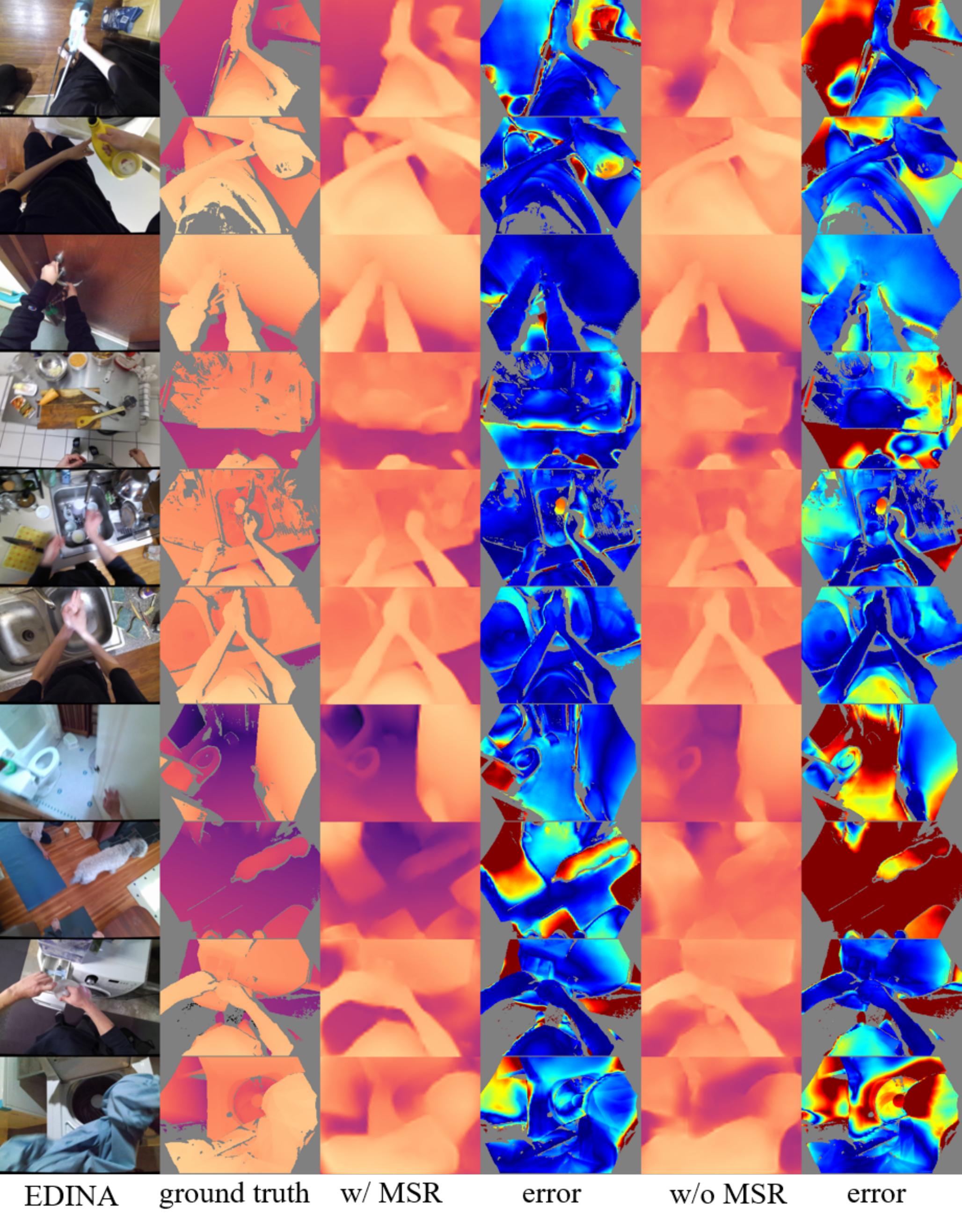}
		\centering
		\caption{Qualitative results for depth prediction on EDINA dataset. From left to right: (1)~RGB image, (2)~ground truth depths,  (3)~depths prediction using PFPN+MSR and its error (the hotter the higher error), and (4)~depths prediction using PFPN and its error.} \label{fig:qualitative_depth_error}
		\label{fig:result}
	\end{figure*}

\begin{figure*}[t]
		\includegraphics[width=0.95\textwidth]{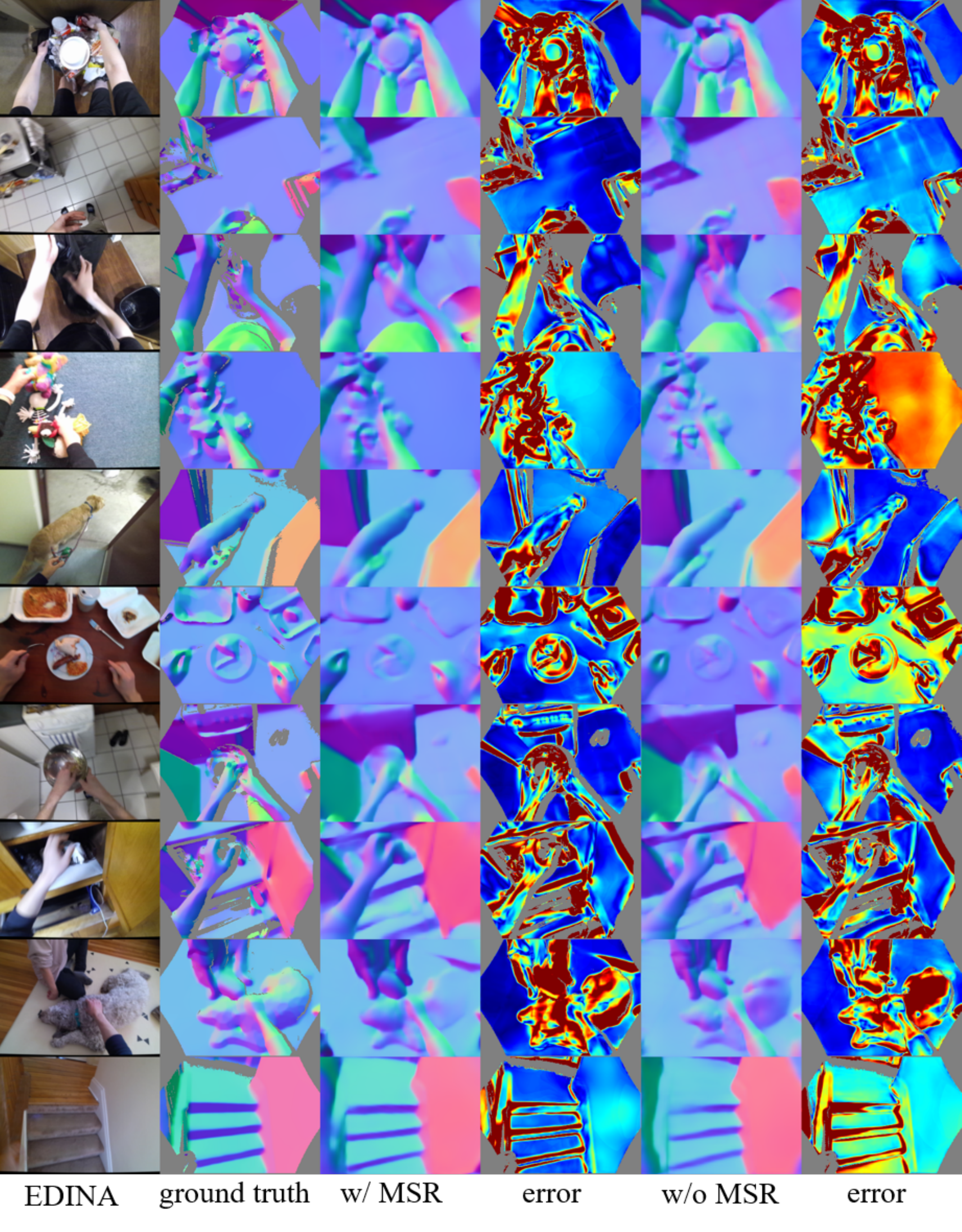}
		\centering
		\caption{Qualitative results for surface normal prediction on EDINA dataset. From left to right: (1)~RGB image, (2)~ground truth surface normals,  (3)~surface normals prediction using PFPN+MSR and its error (the hotter the higher error), and (4)~surface normals prediction using PFPN and its error.} \label{fig:qualitative_sn_error}
		\label{fig:result}
	\end{figure*}

	\noindent\textbf{Qualitative Comparison} We show the qualitative comparison on depths and surface normals estimation with and without the multimodal spatial rectifier on Figure~\ref{fig:qualitative_depth_error} and Figure~\ref{fig:qualitative_sn_error}, respectively.

\begin{figure*}[t]
		\includegraphics[width=0.95\textwidth]{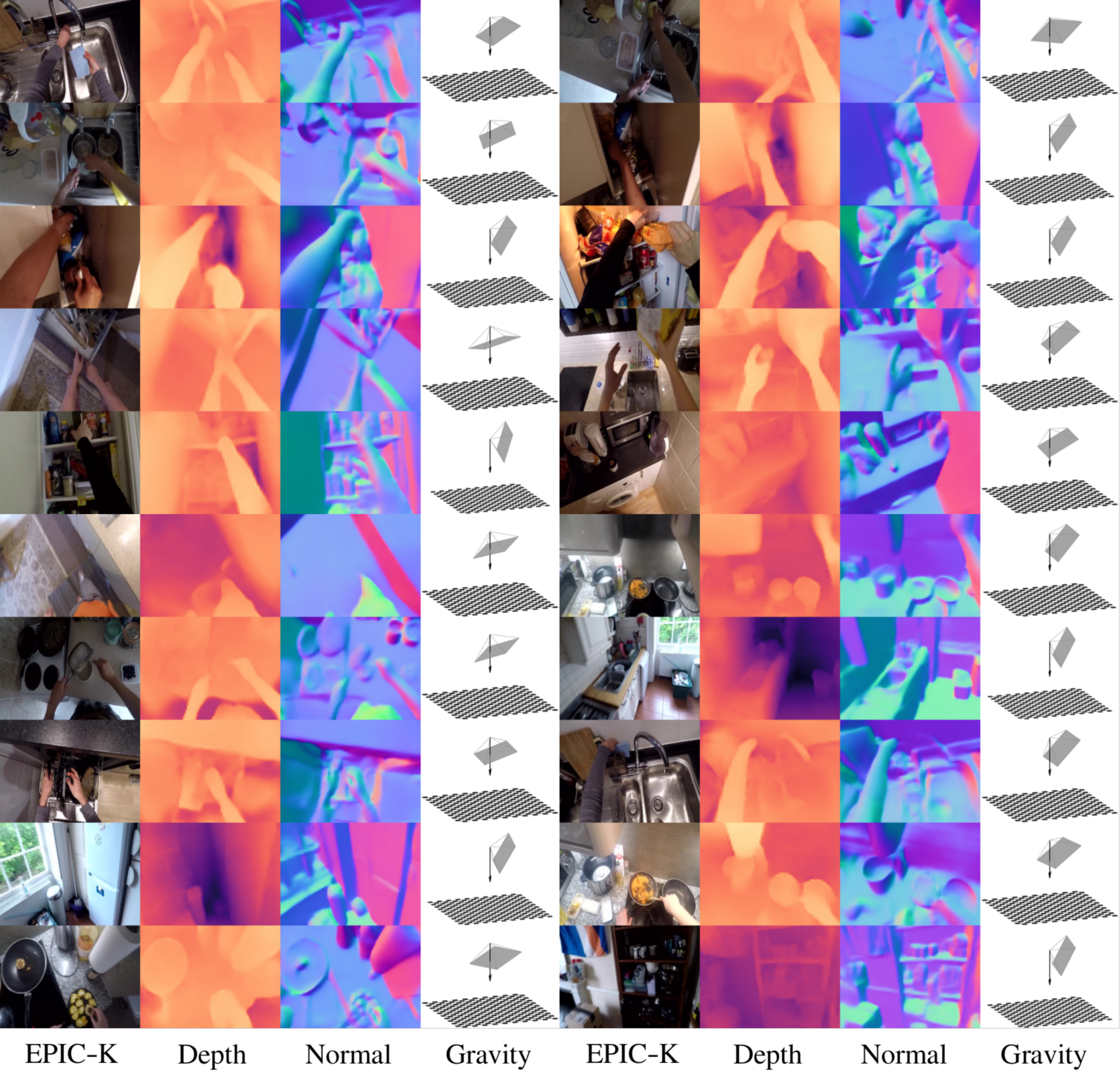}
		\centering
		\caption{Qualitative results for depth, surface normal, and gravity prediction on EPIC-KITCHENS dataset. In each column, from left to right: (1)~RGB image, (2)~depth prediction using PFPN+MSR, (3)~surface normals prediction using PFPN+MSR, and (4)~gravity prediction.} 
		\label{fig:qualitative_epick}
	\end{figure*}

	\noindent\textbf{Qualitative Results on EPIC-KITCHENS} Figure~\ref{fig:qualitative_epick} illustrates the depths, surface normals and gravity prediction on the EPIC-KITCHENS dataset using our multimodal spatial rectifier trained on the ScanNet and our EDINA dataset.
	
\end{document}